\documentclass[wcp]{jmlr}

\usepackage{longtable}

\usepackage{booktabs}
\usepackage{natbib}
\usepackage{graphicx}
\usepackage{float}
\usepackage{subfig}
\usepackage{lipsum}
\usepackage{tabularx}
\usepackage{caption}

\restylefloat{figure}  

\makeatletter
\let\Ginclude@graphics\@org@Ginclude@graphics 
\makeatother

\jmlrvolume{189}
\jmlryear{2022}
\jmlrworkshop{ACML 2022}

\title[Evaluating the Perceived Safety of Urban City via MEDIRL]{Evaluating the Perceived Safety of Urban City via Maximum Entropy Deep Inverse Reinforcement Learning}

  \author{\Name{Yaxuan Wang} \Email{yizhixiaozhu24@gmail.com}\\
  \AND
  \Name{Zhixin Zeng} \Email{2019141440252@stu.scu.edu.cn}\\
  \AND 
  \Name{Qijun Zhao} \Email{qjzhao@scu.edu.cn}\\ 
  \addr Sichuan University, Chengdu, China
 }

\editors{Emtiyaz Khan and Mehmet G\"{o}nen}

\begin{document}

\maketitle

\begin{abstract}
Inspired by expert evaluation policy for urban perception, we proposed a novel inverse reinforcement learning (IRL) based framework for predicting urban safety and recovering the corresponding reward function. We also presented a scalable state representation method to model the prediction problem as a Markov decision process (MDP) and use reinforcement learning (RL) to solve the problem. Additionally, we built a dataset called SmallCity based on the crowdsourcing method to conduct the research. As far as we know, this is the first time the IRL approach has been introduced to the urban safety perception and planning field to help experts quantitatively analyze perceptual features. Our results showed that IRL has promising prospects in this field. We will later open-source the crowdsourcing data collection site and the model proposed in this paper.
\end{abstract}

\section{Introduction}
Urban vitality is receiving increasing attention. As the gathering place for residents to live, the street view profoundly affects the physical and mental health of residents in urban activities~\cite{cheng2017use}. In developed countries, the visual perception of streets receives considerable attention in the urban planning and construction field~\cite{ozkan2014assessment}. More specifically, the residents' perceptions about the safety of the urban scene will influence city layout, as it is a prerequisite for happiness, health, and high quality of life~\cite{troy2012relationship}. The visual perceptions of urban spaces affect the psychological states of their inhabitants and can induce adverse social outcomes~\cite{porzi2015predicting}.
 
With the development of street view technology~\cite{runge2016no} and machine learning (ML), experts have started to use large-scale data for urban safety perception analysis~\cite{dubey2016deep,keizer2008spreading}. Street view images have become essential supporting data for research~\cite{cheng2017use}. \cite{salesses2013collaborative} first proposed that visual attributes of street view can affect safety perception. They also stated that online street view images could be used to create reproducible quantitative measures of urban perception and the inequality of different cities. Subsequent work has investigated urban safety perception using support vector regression \cite{naik2014streetscore}, deep convolutional neural networks~\cite{dubey2016deep}, or multi-instance regression \cite{liu2017place}. \cite{cheng2017use,quercia2014aesthetic} proposed visual features which affect safety perception.

Most of the current research on urban safety perception uses visual features and designs algorithms to obtain the safety level of the street view~\cite{dubey2016deep,naik2014streetscore} and its impact on perception \cite{zhou2021using}. However, most of these methods rely on large-scale data with annotations. In addition, they lack interpretability in quantifying the impact of features on urban safety perception \cite{zhou2021using}. Still, no systematic and comprehensive model based on visual-semantic features \cite{frome2013devise,karpathy2015deep} is available.

To solve the above problems, we introduce decision-making \cite{mnih2015human,you2019advanced} to the field of urban perception for the first time. Specifically, we introduced the Markov decision process (MDP) to model the interaction between the ``expert" and the street view images. Then, we use RL methods to solve the MDP problem and obtain the optimal policy. The agent is assumed to be a decision-maker who interacts with the environment. It receives a reward and a representation of the environment's state at each time step and exerts an action, i.e., the safety evaluation on the environment \cite{you2019advanced}. The reward function is crucial for most RL applications as it can provide a guideline for optimization. However, current RL-based methods rely on manually-defined reward functions, which cannot adapt to dynamic and noisy environments. Besides, they generally use task-specific reward functions that sacrifice generalization ability \cite{chen2021generative}.
 
IRL algorithms can address these problems by learning to imitate the prediction policy produced by an ``expert" \cite{yang2020predicting,baee2021medirl}. IRL is an advanced form of imitation learning \cite{wulfmeier2015maximum} that enables a learning agent to acquire skills from expert demonstrations. This way, the reward function and optimal urban safety perception prediction policy can be obtained. In this paper, our framework obtains rich visual semantic information (environment) by perceptually resolving scenes and proposes a novel state representation method B-VCB. With the help of street view image input, objective safety annotations \cite{liu2017place}, and B-VCB, this framework improves the accuracy and applicability of urban safety perception prediction. Besides, after obtaining the reward function, we can quantitatively analyze the effect of each visual feature.
 
Additionally, we introduce SmallCity, a new dataset used in our study. This dataset should be helpful for urban planners, economists, and social scientists looking to analyze urban perception's social and economic consequences \cite{naik2014streetscore}. 
\begin{figure}[htp]
\begin{center}
\includegraphics[width=0.85\textwidth]{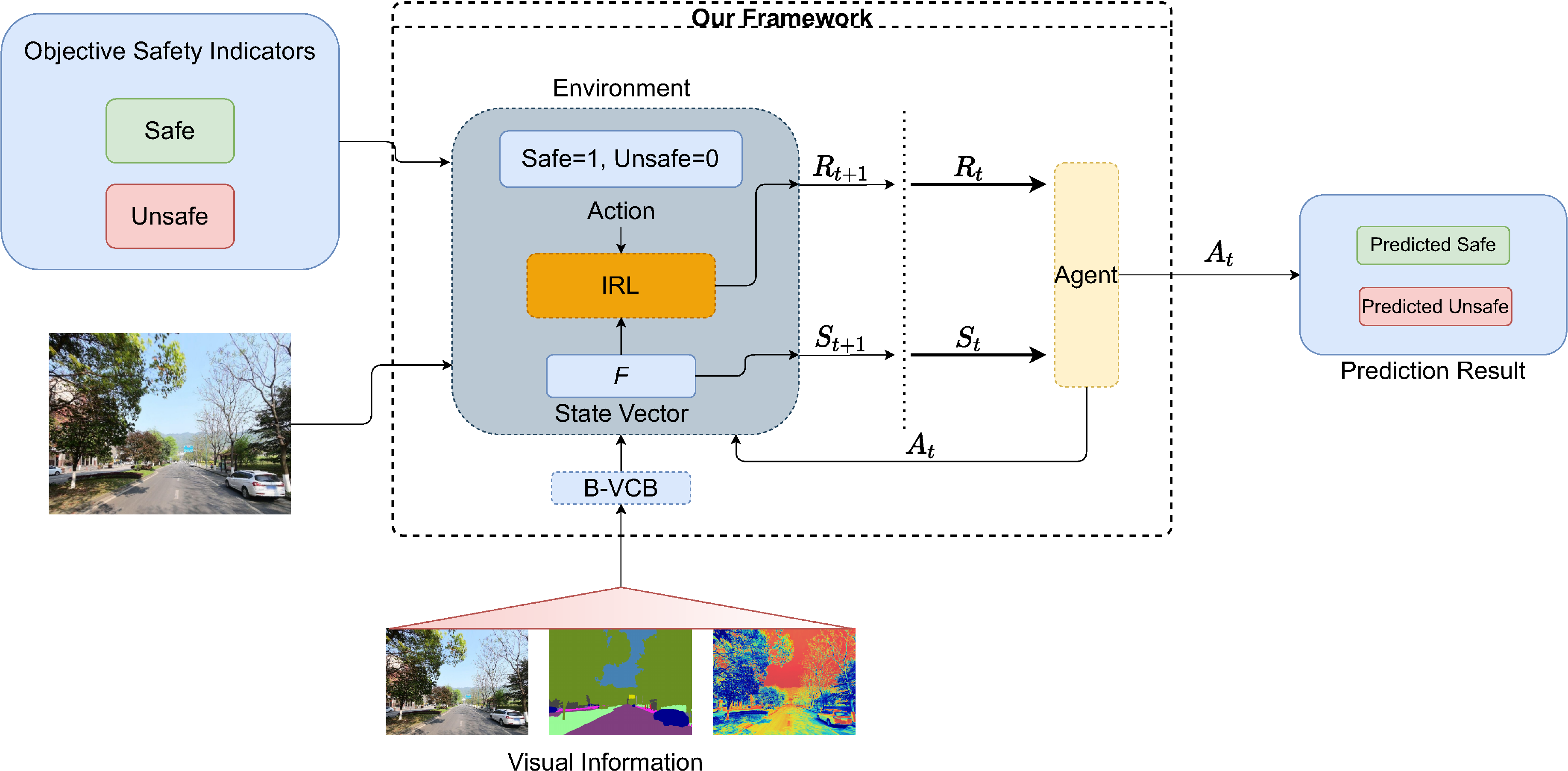}
\caption{Given a street view image and corresponding objective safety indicator inputs, our framework learns to model the prediction policy as states and actions ($S_t$, $A_t$). Then we recover the reward function using inverse reinforcement learning and solve the MDP problem by reinforcement learning.}\label{fig:1}
\end{center}
\end{figure}

Our contributions can be summarized as follows:
\begin{itemize}
  \item [1)] 
  We propose a novel framework that applies maximum entropy deep inverse reinforcement learning (MEDIRL) for predicting urban safety perception and recovering the reward function (Fig.~\ref{fig:1}). 
  \item [2)]
  We present a scalable state representation method to represent the visual features.
  \item [3)]
  We build the SmallCity dataset to investigate the factors influencing safety perception and their effects.
\end{itemize}
\section{Related work}
\subsection{Urban Safety Prediction}
With easy access to large-scale annotated street view datasets and advanced data-driven ML methods, the prediction of urban safety perception has received significant interest \cite{porzi2015predicting,dubey2016deep,liu2017place,zhou2021using}.

Previous studies \cite{naik2014streetscore,arietta2014city} applied support vector regression to predict the safety of the street view based on selected features. Researchers also studied the use of CNN to predict the safety of the street view by analyzing image characteristics~\cite{dubey2016deep,liu2017place}. Conversely, fewer works tried to propose a computational model for explaining visual patterns associated with safety perception.~\cite{porzi2015predicting} presented an approach for predicting the safety of images using CNN. It could automatically discover mid-level visual features correlated with urban perception. More recently, some works explored the relationship between visual semantic features and the occurrence of criminal activities in the context of urban perception \cite{jing2021assessing,zhou2021using}.

The above studies of urban safety perception are based on crowdsourcing annotations or data from the government's criminal records. These methods require a large-scale dataset to obtain precise predictions, mainly rely on feature selection and lack semantic level interpretability~\cite{naik2014streetscore,dubey2016deep}. More importantly, the crowdsourcing annotations have unavoidable noise and cannot fully represent the actual perception. 

We are interested in simulating the experts' decision-making process using crucial visual semantic features to obtain a safe evaluation policy for street view images. A quantitative analysis of the impact of images' visual features is possible based on the evaluation results.

\subsection{Inverse Reinforcement Learning}
Our IRL approach follows the methods in which problems are modeled as a sequential decision process to solve \cite{yang2020predicting,baee2021medirl}. Classical IRL considers expert demonstrations, i.e., a finite set of trajectories and knowledge of the environment to find the expert's potential reward function \cite{arora2021survey}. Some IRL algorithms \cite{wulfmeier2015maximum,yang2020predicting} learned an optimal policy and the corresponding reward function simultaneously. The research concentrated on the planning problem of autonomous vehicles in traffic \cite{you2019advanced} is closest to our research as it proposed a Single-Step Joint MEDIRL. However, our method does not need to generate demonstrations through RL and learn the corresponding policy. Instead, we use crowdsourcing to collect the images and safety annotations as demonstrations and recover the intrinsic reward functions \cite{zheng2018learning} induced by experts' evaluation strategies. We used the MEDIRL method \cite{you2019advanced} that can take the raw image as inputs and enables the model to handle experts' sub-optimal policies and seemingly stochastic behaviors to achieve our goal.

\subsection{Street View Dataset}

Crowdsourcing websites have been used to collect large-scale data \cite{quercia2014aesthetic}. The website designed in \cite{salesses2013collaborative} used questions (e.g., ``Which place looks safer?") and images from Google Street View to obtain comparison pairs. Furthermore, the safety score of each image will be calculated using TrueSkill \cite{herbrich2006trueskill} or RSS-CNN~\cite{dubey2016deep}.

\begin{figure}[htbp]
\begin{center}
\includegraphics[width=0.85\textwidth]{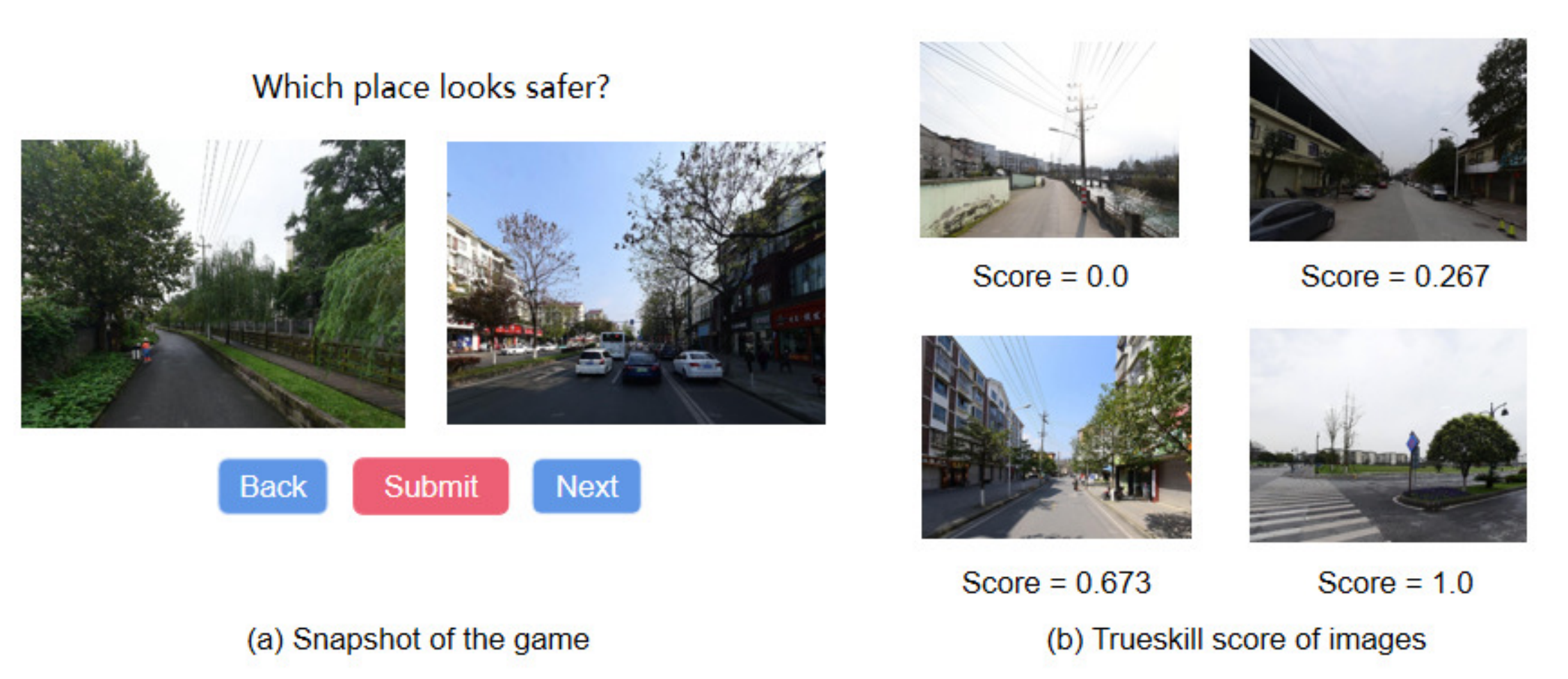}
\caption{The website used to get pairwise comparisons of urban safety perception (a). The relative safety score of each image obtained using TrueSkill (range between [0,1]) (b).}\label{fig:2}
\end{center}
\end{figure}

We followed \cite{salesses2013collaborative} to design our website (Fig.~\ref{fig:2}) and build the dataset used in research based on Dujiangyan City's street view.
\section{The SmallCity Dataset}

We proposed the SmallCity dataset, which consists of 522 geographically averaged street view images of Dujiangyan City. Each of these images has corresponding semantic features (e.g., FOV), perceptual features (e.g., visual entropy), quantitative features (e.g., number of cars or pedestrians), and corresponding objective safety perception annotation. Our study considers the safety annotation as experts' predictions.

\subsection{Obtaining street view images}
45,412 images and corresponding latitude-longitude data were initially obtained using the Baidu Map API. To reflect the overall appearance of Dujiangyan City, 522 geographically averaged images were used to build the SmallCity dataset.

\subsection{Crowdsourcing Website}
Following \cite{salesses2013collaborative}, we designed a website (Fig.~\ref{fig:2}) to collect pairwise comparisons. Our
website presents two images randomly to obtain comparison
pairs, which is more acceptable and accurate when getting
safety scores~\cite{porzi2015predicting,stewart2005absolute,bijmolt1995effects}. To be more specific, we showed users a randomly-chosen pair of images side by side and asked them to choose one in response to the question: \textbf{``Which place can give you a safer impression?"}. Information such as the user's age, gender, and the location was also collected. It is worth noting that there is no significant cultural bias in individual preferences for urban appearance.

\subsection{Image Feature Encoder}
Unlike most previous studies that focused on the general image features \cite{naik2014streetscore,dubey2016deep}, we addressed the semantic visual features \cite{shen2017streetvizor}. We applied the SOTA semantic segmentation model SegFormer \cite{xie2021segformer} on the SmallCity dataset and obtained inference results, i.e., the semantic results, which show approximate contours of each object that appeared on the image. The cover ratio of each feature, i.e., FOV \cite{cheng2017use}, can be calculated.

The definition of street view variables for state representation is shown in Table \ref{table:tab1}. Here we used greenery \cite{cheng2017use} to represent the vegetation and terrain. We selected variables that significantly impact perception to investigate urban safety perception based on previous qualitative studies \cite{zhang2018representing}. Considering \cite{zhou2021using} indicates that wire poles make notably significant and negative impacts on the occurrence of criminal activities, we used the binary value to represent the existence of wire poles. The definitions of our street view variables are shown in Table \ref{table:tab1}.

Additionally, using TrueSkill, we obtained the relative ranking of each image. Furthermore, we calculated the average of the ranking scores and used them as the threshold in combination with expert ratings to derive the objective safety annotations.

\begin{table}[htbp] 
    \begin{center}
    \centering
	\caption{Specific definitions of street view variables.}
    \begin{tabularx}{\textwidth}{cX}
 		\toprule
 	 	\textbf{street view Variables} & \textbf{Definitions} \\ 
  		\midrule 
 		Greenery & The cover ratio of terrain and vegetation elements in the field of view  \\ 
 		Sky & The cover ratio of sky element in the field of view  \\ 
 		Wall & The cover ratio of wall element in the field of view  \\ 
 		Fence & The cover ratio of fence element in the field of view  \\ 
 		Sidewalk & The cover ratio of sidewalk element in the field of view  \\ 
 		Electrical Wire & The value of this feature is “1” if the wire pole element is in the field of view and "0" otherwise  \\ 
 		Visual Entropy & Measure the amount of information within an image and the sensitivity of the human visual system to the image~\cite{cheng2017use}  \\ 
 		Car Number & The number of cars in the street view  \\ 
  \bottomrule 
  \end{tabularx}\label{table:tab1}
 \end{center}
\end{table}

\section{Method}

We first modeled the urban safety perception prediction problem as an MDP process \cite{bellman1957markovian}. Then we proposed an IRL framework (Fig.~\ref{fig:1}) to model experts' decision process of prediction to obtain the reward function and learn a prediction policy from demonstrated behaviors, i.e., the crowdsourcing annotations.

Targets performing the safety perception task can be considered goal-directed agents, with the predictions being a sequential decision process. At each time step, the agent attends to receive a random street view image and approximates the state based on the corresponding MDP process. The images were represented using B-VCB proposed in this paper (Sec. 4.1). Also, each agent's action depends on the state at the specific time step during the decision process. 
The agent's goal is to maximize the rewards obtained. While it is difficult to measure how much reward is received from these predictions behaviorally, with IRL, the reward can be assumed to be a function of the state and the action. This function can be jointly learned using the MEDIRL method (Sec. 4.2). We can also get the learned expert-like policy.

\subsection{State Representation Modeling}

To model the state of an agent, we propose a novel state representation method called Binary Visual-Contextual-Belief (B-VCB). 

\textbf{Contextual belief} \cite{yang2020predicting}. It represents a person’s gross “what” and “where” understanding of a scene. It means we can assume people have an internal scene parser that takes an image input and generates belief maps based on all the objects and background classes in that person’s knowledge structure when evaluating an image. Different from \cite{yang2020predicting}, to capture more useful perceptual features and form better beliefs, we use SegFormer for image segmentation and select specific visual features, as mentioned in 3.3.

\textbf{Binary representation} 
Considering the impact of features on safety perception tends to be linear and monotonic \cite{zhou2021using}, and the value ranges of different features are different, we use the threshold segmentation method to divide those greater than and those less than the threshold value into two different states, i.e., 0 or 1. The threshold value is the average of the current feature. Furthermore, "1" and "0" respectively mean the feature represents an unsafer or safer factor.

Following \cite{you2019advanced}, 256 states can be obtained through the MDP process, and each state $s$ is represented as a state vector $F$ of a binary set. It is worth mentioning that more corresponding classes can be obtained when needed besides the street view variables used in our research. For instance, other variables such as Salient Region Saturation \cite{cheng2017use}, traffic lights \cite{zhou2021using}, and the number of people can be considered.
\begin{figure}[htp]
\begin{center}
\includegraphics[width=0.6\textwidth]{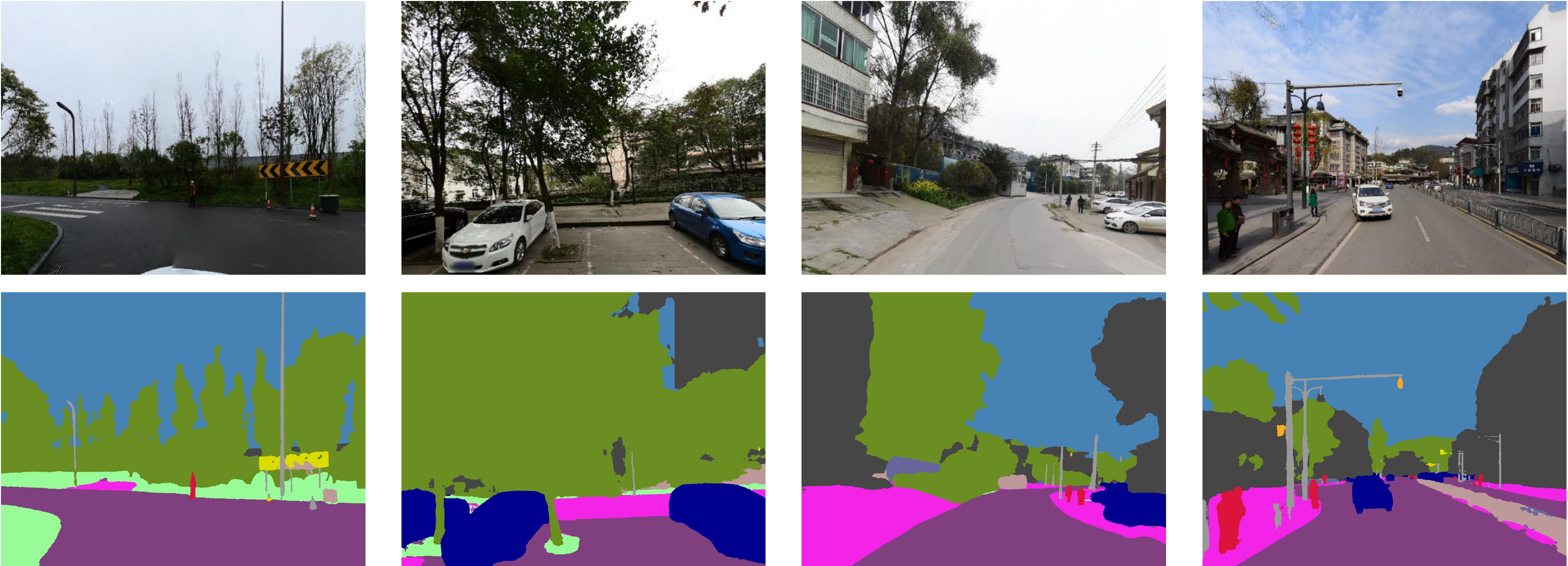}
\caption{Example of inference results of the SmallCity dataset}\label{fig:4}
\end{center}
\end{figure}
\subsection{Reward and Policy Learning}
We use MEDIRL to learn the reward function and urban safety perception prediction policies in a small dataset. MEDIRL assumes the reward is a function of the state and the action, and this reward function can be jointly learned using the imitation policy \cite{baee2021medirl}.

\textbf{MDP Notations} 
In our work, $S$ is a set of possible states; $A$ is a set of available actions, which is given by $A$ = \{``safe", ``unsafe"\}; $F$ is the set of the visual features for a state; ${\tau}$ is the set of images and corresponding demonstrations. Specifically, we use Single-Step Joint MEDIRL \cite{you2019advanced} to recover the unknown reward function $R$ from the set of demonstrations $
\varGamma _{\tau}=\left\{ \xi _{\tau}^{i},\ i=1,...,N_{\tau} \right\}$, which is generated using the crowdsourcing method, and obtain the optimal policy. The element $\xi _{\tau}^{i}\ =\ \left\{ \left( s_i,a_i \right) \right\} $ satisfies the following conditions: (1) $\xi _{\tau}^{i}$ starts at the state $\tau \in S$ (2) The length of $\xi _{\tau}^{i}$ is constant $\bigtriangleup T$ for all $\tau \in S$ (Here we have $\bigtriangleup T$ = 1), and (3) There exists a path $\zeta \in D$ such that $\xi _{\tau}^{i}\subseteq \zeta$. The corresponding path space for is denoted as $\varOmega _{\tau}$. We then maximized the entropy of the joint distribution over all $\varOmega _{\tau}$ subject to the constraints from the demonstrations $\varGamma _{\tau}$ following \cite{you2019advanced}.

\textbf{Reward Function} To obtain the prediction policy, we need to reproduce the agent’s preference for specific actions when facing different states. Thus we chose the state-action-reward $R:S\times A\rightarrow \mathbb{R}$ \cite{levine2012continuous}, which takes the action into consideration. 

A widely used approach to design the reward function is to represent it as a linear combination of some manually chosen features \cite{ziebart2008maximum}; \cite{levine2012continuous}; \cite{yang2020predicting}; \cite{baee2021medirl}. 
\begin{equation}
R\left( s,a \right) =\omega ^T\varPhi \left( s,a \right) 
\end{equation}

\begin{figure}[htp]
\begin{center}
\includegraphics[width=1.0\textwidth]{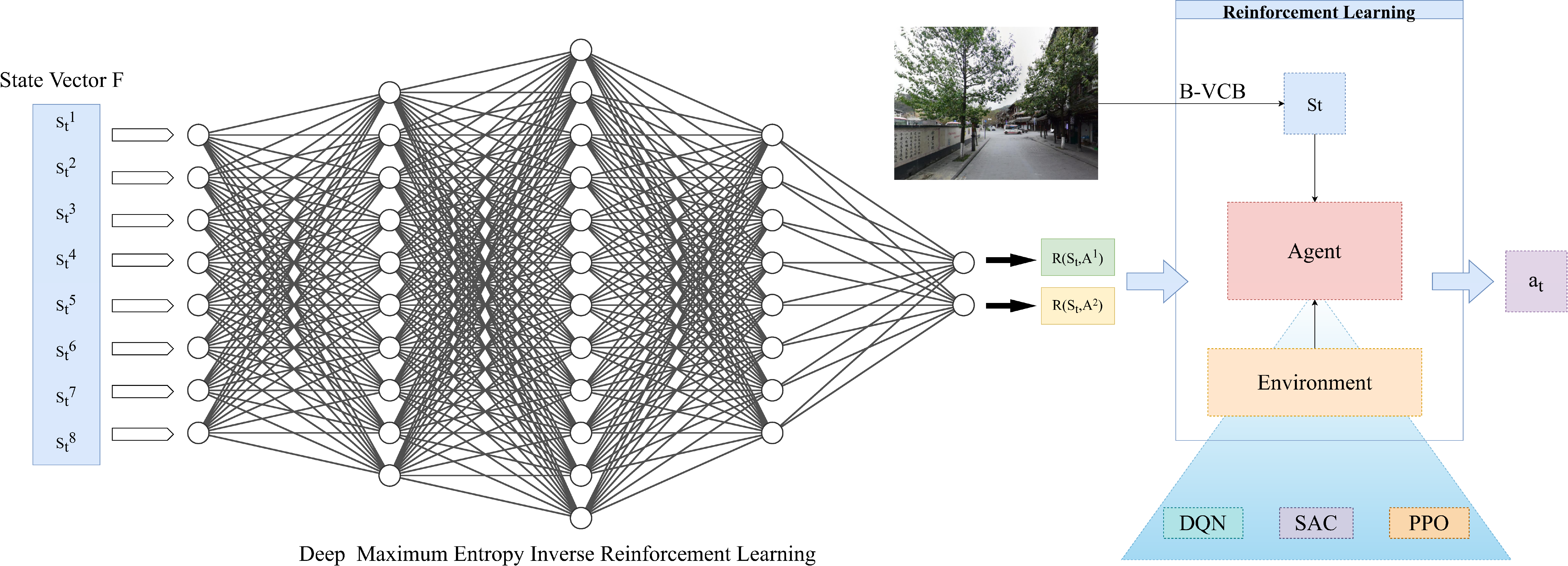}
\caption{Recovering \textbf{reward functions} using DNN. Note that the number of layers and neurons in the hidden layer is adjustable. The DNN produces the rewards considering every action in action set $A$. One then just needs to select the correct output channel according to the current action $a_t$. }\label{fig:5}
\end{center}
\end{figure}

In our work, we use a deep neural network (DNN) that can approximate any function as a parameterized reward function \cite{wulfmeier2015maximum} (Fig.~\ref{fig:5}). These features only depend on the current state $s$, which the feature vector $ F$ can represent. Note that the weight vectors of $\omega$ and the parameter vector $\theta$ are both associated with the network, which is fine-tuned through training. 
\begin{equation}
R\left( \omega ,\theta ;s,a \right) \ =\ \omega ^T\varPhi \left( \theta ;s,a \right) 
\end{equation}

\textbf{Policy Learning} To learn the prediction policies, Single-Step Joint MEDIRL maximizes the joint posterior distribution of demonstrations $\varGamma _{\tau}$, under a given reward structure and of the model parameter, $\omega$ and $\theta$. It corresponds to maximizing the following objective function:
$$
L_{D(\omega, \theta)}=\frac{1}{N} \sum_{\xi \in I} \log P(\xi \mid \omega, \theta)
$$
where the probability of a demonstration $\xi _{\tau}^{i}$, $P\left( \xi |\omega ,\theta \right)$ is calculated following Theorem 4.1 in \cite{ziebart2008maximum}.
We can also obtain the loss gradient and maximum-entropy gradient \cite{you2019advanced}:
\begin{equation}
\frac{\partial \mathbb{L}_{D\left( \omega ,\theta \right)}}{\partial \theta}\ =\ \sum_{s\varepsilon S}{\sum_{a\epsilon A}{\frac{\partial \mathbb{L}_{D\left( \omega ,\theta \right)}}{\partial R\left( \omega ,\theta ;s,a \right)}\ \frac{\partial R\left( \omega ,\theta ;s,a \right)}{\partial \theta}}}\label{four} 
\end{equation}
where $\frac{\partial R\left( \omega ,\theta ;s,a \right)}{\partial \theta}$ can be obtained by backward propagation and $\frac{\partial \mathbb{L}_{D\left( \omega ,\theta \right)}}{\partial R\left( \omega ,\theta ;s,a \right)}\,\,$ is the maximum-entropy gradient:

\begin{equation}
\frac{\partial \mathbb{L}_{D\left( \omega ,\theta \right)}}{\partial R\left( \omega ,\theta ;s,a \right)}\ =\ \mu _D\left( s,a \right) -E\left[ \mu \left( s,a \right) \right]    
\end{equation}
where $\mu _D\left( s,a \right)$ is the expected empirical state–action pair visitation counts over the demonstrations $\varGamma _{\tau}$, and $E\left[ \mu \left( s,a \right) \right]$ is the expected state–action pair visitation counts calculated over $\bigtriangleup T$ steps. In our work, we let $\bigtriangleup T$ = 1 and consider only the one-step action case, so we have:
\begin{equation}
\frac{\partial \mathbb{L}_{D\left( \omega ,\theta \right)}}{\partial R\left( \omega ,\theta ;s,a \right)}\ =\ \pi _D\left( s,a \right) -\pi \left( s,a \right) 
\end{equation}
where $\pi _D\left( s,a \right)$ is the expected empirical policy and $\pi \left( s,a \right)$ is the learned policy.

The IRL method aims to optimize \ref{four} and get the weights $\omega$, $\theta$ of the reward function. After obtaining the reward of an “expert,” we then use reinforcement learning techniques to solve our MDP problem to get the optimal policy $\pi ^*\left( s,a \right) $.

\section{Experiments}
\subsection{Implementation Details}

\textbf{Feature Encoding.} To encode visual information (Sec. 3), we used several semantic segmentation models: PSPNet \cite{zhao2017pyramid}, DeepLab V3+ \cite{chen2018encoder}, Segmenter \cite{strudel2021segmenter} and SegFormer. After comparing the results of these methods, we applied SegFormer in our research (Fig.~\ref{fig:4}) and calculated the FOV of the selected features. Then we encode each image into a state vector $F$.

\textbf{Reward Representation} We only used the state $s_t$ as the input of the DNN and provided the rewards for taking every action in action set $A$. The state vector $F_t$ contains the visual information for urban safety perception. Hence, the DNN requires eight input channels to receive the eight-dimensional state $s_t$. The DNN requires two output channels since there are two actions in $A$. We define a DNN with the numbers of the neurons in each layer given by [8, 32, 16, 32, 2], which are hyperparameters. We used Tanh instead of Sigmoid because it is unbiased at origin and has a more extensive range. 

\textbf{Policy Learning} We used OpenAI Gym to implement our environment and Ray RLlib to implement the RL algorithms: PPO \cite{schulman2017proximal}, SAC \cite{haarnoja2018soft}, and D3QN \cite{van2016deep,wang2016dueling}. 

\textbf{Evaluation Metrics} We used learned behavior accuracy to evaluate the IRL module for reward function recovery \cite{arora2021survey}. To evaluate urban safety perception predictions, we used F1-Score and Area under the ROC curve (AUC) to measure the accuracy of the policy.

\subsection{Results and Discussion}
\begin{table}[htbp] 
    \begin{center}
    \centering
	\caption{Evaluation of performances on SmallCity.} 
    \begin{tabular}{cp{8cm}<{\centering}cp{8cm}<{\centering}cp{8cm}<{\centering}}
 		\toprule 
 	 	\textbf{Method} & \textbf{F1-Score (\%)} & \textbf{AUC} \\ 
  		\midrule 
 		SVM & 61.9\% & 0.610  \\ 
 		FCN & 61.4\% & 0.628  \\ 
 		ResNet-101 & 53.9\% & 0.532  \\ 
 		ViT & 66.7\% & 0.618  \\ 
 		\textbf{Ours} & \textbf{69.3\%} & \textbf{0.709}  \\ 
        \bottomrule 
     \end{tabular}\label{table:tab2}
 \end{center}
\end{table}
\begin{table}[htbp] 
    \begin{center}
    \centering
	\caption{Performance comparison of urban perceived safety prediction on SmallCity using different RL methods and different reward functions. }  
    \begin{tabular}{cp{8cm}<{\centering}cp{8cm}<{\centering}}
 		\toprule 
 	 	\textbf{Method} & \textbf{Learned Behavior Accuracy(\%)}  \\ 
  		\midrule 
 		IRL + PPO & 63.6   \\ 
 		IRL + SAC & 70.5   \\ 
 		\textbf{IRL + D3QN} & \textbf{71.9}  \\ 
 		Expert Reward + D3QN & 68.5  \\ 
        \bottomrule 
     \end{tabular}\label{table:tab3}
 \end{center}
\end{table}

We compare the proposed framework with traditional ML and deep learning (DL) methods. For ML methods, we applied SVM \cite{steinwart2008support} by using the street view variables selected to evaluate the safety perception. While for the DL approach, we used fully connected neural networks (FCNNs) to obtain predictions based on variables. We also used ResNet-101 \cite{he2016deep} and ViT \cite{dosovitskiy2020image} for safety perception evaluation based on the street view images. Since these methods are typical algorithms for solving this problem and the models are accessible, they are used as experimental comparisons.

Table \ref{table:tab2} provides the quantitative evaluation results of our framework and shows that our IRL+RL framework outperforms other methods on the SmallCity dataset in terms of the F1-Score and AUC. Most significantly, our approach can effectively predict urban safety perception. It demonstrates that our method is advantageous for accurately anticipating if a street view image is safe or not. ViT’s result being 3\% lower than ours in terms of the F1-Score is the inherent problem of supervised learning, which lacks high-level interpretability. The poorer result of ResNet-101 is related to the overfitting problem. IRL+D3QN achieves the best performance in our study, as shown in Table 2. IRL+SAC 
and IRL+D3QN, which used IRL to recover reward functions, achieved higher learned behavior accuracy than the method using an expert-designed function. The results demonstrate the effectiveness of the IRL method.

Table \ref{table:tab3} shows the comparison of different RL algorithms when solving the MDP problem. We also compared the IRL recovered reward function with the expert design reward function. 


\begin{figure}[ht]
\subfloat[Total expert rewards in the training process.]{
    \label{pica}
    \includegraphics[width=6.8cm]{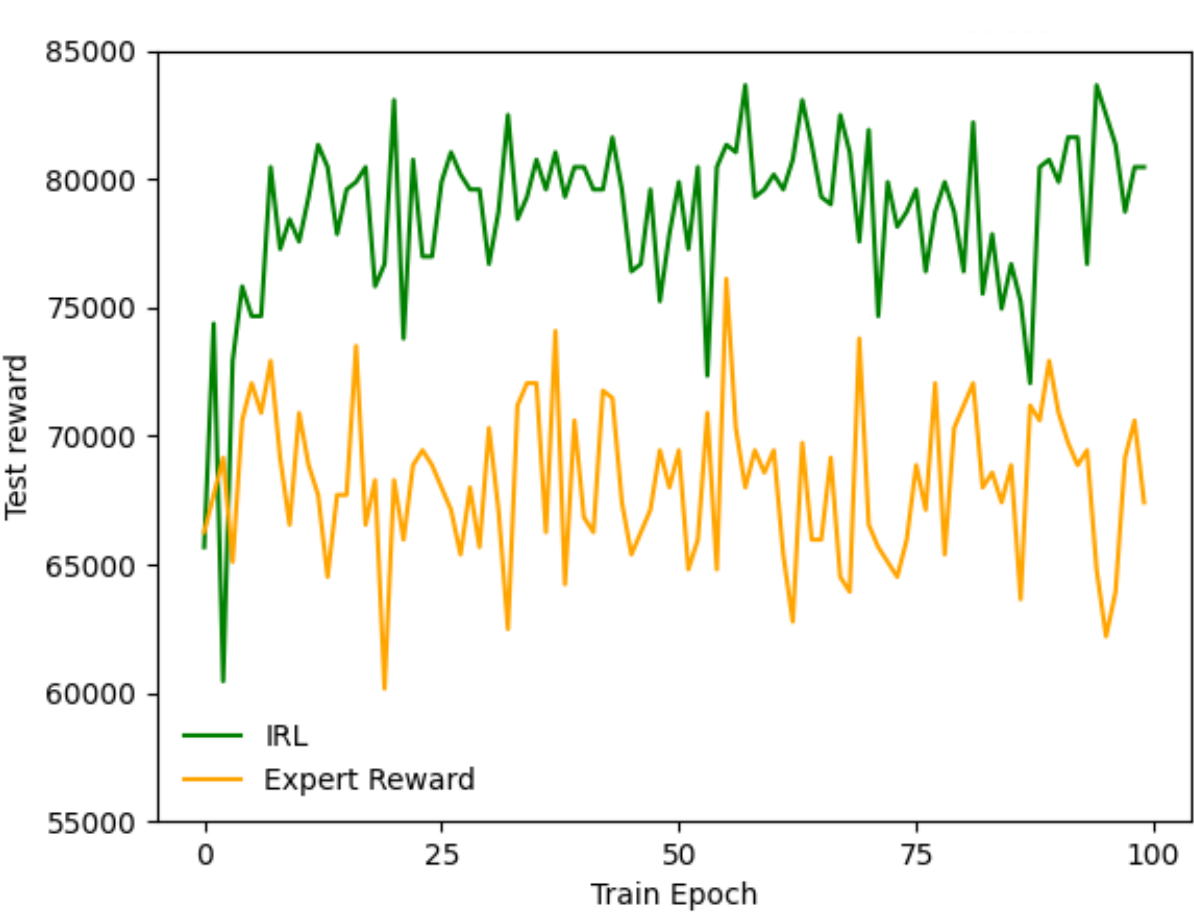}
}
\subfloat[The reward value of each feature on one image.]{
    \label{picb}
    \includegraphics[width=9cm]{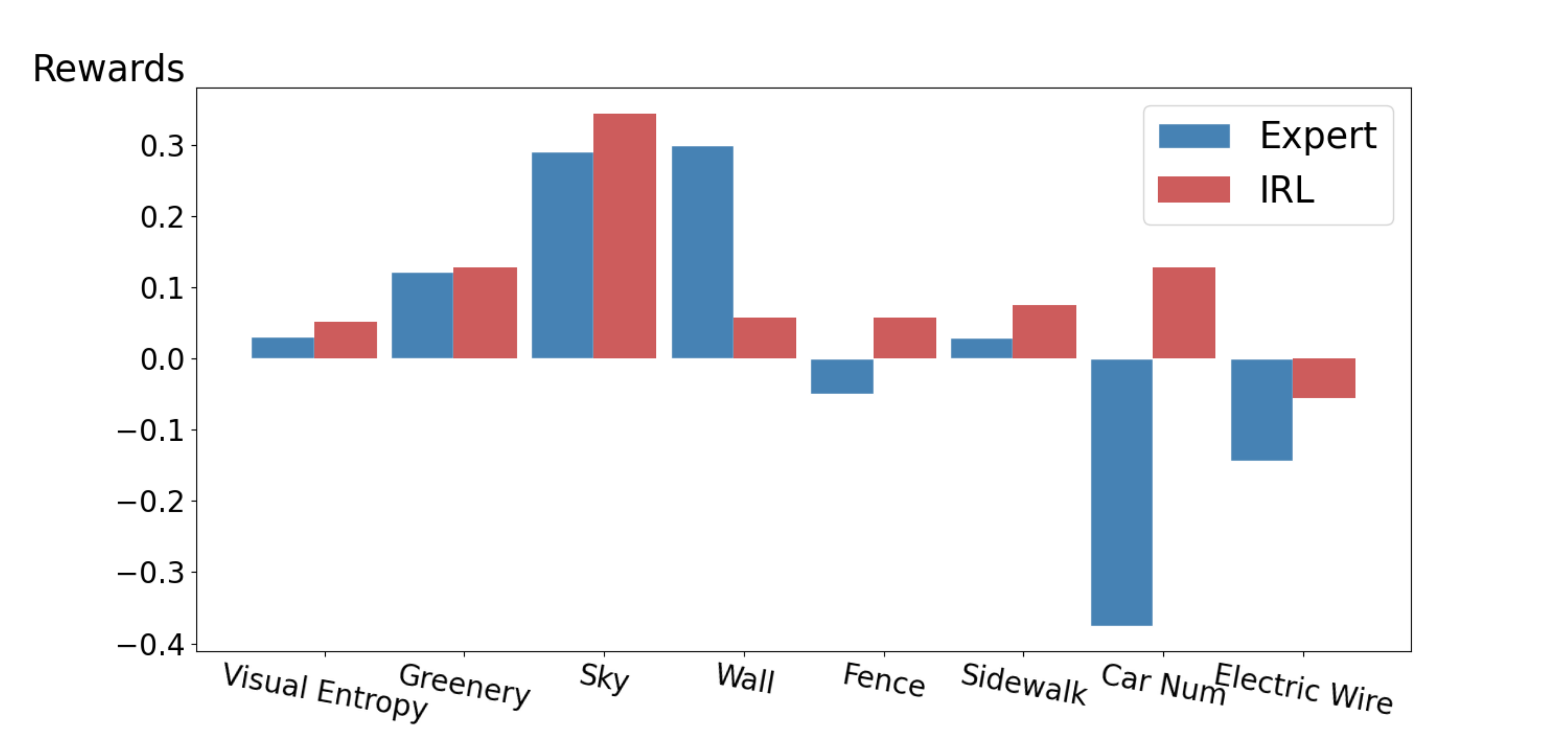}
}
\label{fig5}
\caption{The reward curves during learning process under different
reward functions (a). The reward value of recovered reward (IRL) and expert
reward corresponds to each feature (b).}
\end{figure}


We calculated the ``total expert rewards" encountered when traversing trajectories generated using policies under IRL learned reward or expert reward for evaluation, as shown in Fig.5 \subref{pica}. That means we use IRL learned and expert rewards separately during training. In the evaluation, we only used the expert reward metric to address the problem of the value range. We carefully compared the results of IRL and expert reward functions on the same image. Different reward function assigns each feature a score. It shows that IRL recovered reward function can provide better training signals than supervised learning and achieve the increasing reward. 

When designing the reward function, the experts analyzed the selected features and determined threshold values of ``reward" or ``penalty." Each expert also gave each image a final prediction result, i.e., safe or unsafe. We obtained the expert reward by considering the experts' evaluation strategy. Fig.5 \subref{picb} shows reward values of each feature when the image's predictions are consistent with the label, i.e., safe. The expert-designed reward function is based on empirical evidence, while the recovered reward function is more consistent with the crowdsourcing annotations. The former is a simple function, while the latter is a multilayer neural network. Therefore, the learned reward function gives some features a different interpretation from the expert reward.

\subsection{Ablation Study}
We use the IRL method to recover the reward function and mimic the human intrinsic perception patterns. More specifically, for a decision pair, we can obtain the reward function value of the decision behavior, demonstrating the degree of fitting the expert evaluation strategy. A more considerable reward value means making the judgment for one picture more constant with the expert's evaluation strategies. By changing the value of each feature dimension (e.g., by setting the value of uninteresting features to 0), we can focus on certain street-level features and learn the relationship between the input features and the safety perception by the recovered reward function.

\begin{table}[htbp] 
    \begin{center}
    \centering
	\caption{Ablation study of the proposed state representation method
Binary Visual-Contextual-Belief.} 
    \begin{tabular}{cp{3cm}<{\centering}cp{3cm}<{\centering}cp{3cm}<{\centering}}
 		\toprule 
 	 	\textbf{B-VCB Settings} & \textbf{F1-Score (\%)} & \textbf{AUC}& \textbf{Acc (\%)} \\ 
  		\midrule 
 		B-VCB full & 69.3\% & 0.700 & 71.9\% \\
 		B-VCB w/o Greenery & 61.6\% & 0.632 & 63.3\% \\
 		B-VCB w/o Car Number & 66.5\% & 0.662 & 66.1\% \\
 		B-VCB w/o Electrical Wire & 68.4\% & 0.687 & 69.1\% \\
        \bottomrule 
     \end{tabular}\label{table:ablation}
 \end{center}
 \end{table}
 In Table \ref{table:ablation}, we report the results of ablation studies with the SmallCity dataset. The first row shows the complete B-VCB + IRL + D3QN method results. In the second row, we remove the visual feature Greenery in B-VCB and keep other conditions unchanged. Other rows have declined on the three evaluation metrics compared with our full model(the first row). It proves the effectiveness of our B-VCB method.
 
Our IRL method is more effective than conventional supervised learning because of the combination of B-VCB and IRL. The B-VCB method can simulate the evaluation pattern of experts and extract perceptual features, which are more robust and targeted. Also, the recovered reward function can provide more training signals in obtaining the optimal policy. Our framework can get a policy more similar to experts’ patterns.


\section{Conclusions}
We proposed a novel IRL-based framework for predicting the perceptual safety of street view images in urban perception situations. We also presented a new state representation method of street view images. Our framework effectively recovers the reward function and learns the optimal policies for safety perception; based on this, an urban planner can learn the impact of different semantic visual features on perceptual safety from the reward function. To facilitate our study, we built the SmallCity dataset. We investigate the effectiveness of our method through experimental evaluation of different methods based on the SmallCity. Results show that our method outperforms existing models and indicate that the IRL method has promising prospects in related fields.

\acks{This work was supported in part by the top-notched student program of Sichuan University.}
\bibliography{acml22}

\end{document}